\documentclass{article}

 \usepackage[dblblindworkshop, final]{neurips_2025}

\workshoptitle{Data on the Brain \& Mind}

\usepackage[utf8]{inputenc} 
\usepackage[T1]{fontenc}    
\usepackage{hyperref}       
\usepackage{url}            
\usepackage{booktabs}       
\usepackage{amsfonts}       
\usepackage{nicefrac}       
\usepackage{microtype}      
\usepackage{xcolor}         
\usepackage{graphicx} 
\usepackage{flafter} 
\usepackage{placeins}   
\usepackage{float}      
\usepackage{amsmath}

\title{Massively Parallel Imitation Learning of Mouse Forelimb Musculoskeletal Reaching Dynamics}

\author{%
  Eric Leonardis\thanks{Contact the primary author at ejleonardis@salk.edu} \\
  Salk Institute for Biological Studies\\
  La Jolla, CA 92037\\
  \texttt{eleonardis@salk.edu} \\
  \And
  Akira Nagamori \\
  Salk Institute for Biological Studies\\
  La Jolla, CA 92037\\
  \texttt{anagamori@salk.edu} \\
  \And
  Ayesha Thanawalla\\
  Salk Institute for Biological Studies\\
  La Jolla, CA 92037\\
  \texttt{ghsaunders22@gmail.com} \\
  \And
  Yuanjia Yang\\
  Salk Institute for Biological Studies\\
  La Jolla, CA 92037\\
  \texttt{scyang@salk.edu} \\
  \And
  Joshua Park\\
  Salk Institute for Biological Studies\\
  La Jolla, CA 92037\\
  \texttt{jjp003@ucsd.edu} \\
  \And
  Hutton Saunders\\
  Salk Institute for Biological Studies\\
  La Jolla, CA 92037\\
  \texttt{ghsaunders22@gmail.com} \\
  \And
  Eiman Azim \\
  Salk Institute for Biological Studies\\
  La Jolla, CA 92037\\
  \texttt{eazim@salk.edu} \\
  \And
  Talmo D. Pereira \\
  Salk Institute for Biological Studies\\
  La Jolla, CA 92037\\
  \texttt{talmo@salk.edu} \\
}

\begin{document}

\maketitle

\begin{abstract}
  The brain has evolved to effectively control the body, and in order to understand the relationship we need to model the sensorimotor transformations underlying embodied control. As part of a coordinated effort, we are developing a general-purpose platform for data-driven simulation modeling high fidelity behavioral dynamics, biomechanics, and neural circuit architectures underlying embodied control. We present a pipeline for taking kinematics data from the neuroscience lab and creating a pipeline for recapitulating those natural movements in physics simulation.  We implement an imitation learning framework to simulate a dexterous forelimb reaching task with a musculoskeletal model in the Mujoco physics environment. The imitation learning model is currently training at more than 1 million training steps per second due to GPU acceleration with JAX and Mujoco-MJX. We present results that indicate that adding naturalistic constraints on control magnitude lead to simulated muscle activity that better predicts real EMG signals. This work provides evidence to suggest that control constraints are critical to modeling biological movement control.
\end{abstract}

\section{Introduction}
The brain has evolved to effectively control the body, and in order to understand the relationship it is necessary to model the sensorimotor transformations underlying embodied control. In the motor control literature, it is common to infer neural mechanisms of movement control only from observed movement kinematics without accounting for the underlying dynamics of musculoskeletal systems and their interactions with external environments. Moreover, recent work has shown that using physics-constrained imitation learning to reproduce experimentally observed motor behavior can predict real brain activity [1]. However, it is difficult to accurately measure dynamics from experimental data outside of very well controlled and adequately equipped environment, which may not be feasible for certain animals such as mice. Therefore, simulating neuromechanical interactions may provide a critical step toward understanding embodied control as it can allow us to quantify the landscape of available solutions [2]. 

As part of a coordinated effort to address this gap, we are developing a general-purpose platform, MIMIC-MJX, for behavior-driven simulation modeling high fidelity behavioral dynamics, biomechanics, and neural circuit architectures underlying embodied control [3].  This open source and publicly available library is made of two parts -- 1) STAC-MJX which registers pose estimation data to the biomechanical model to generate reference inverse kinematics and 2) TRACK-MJX which uses imitation learning to reproduce the motion capture in a physics environment. MIMIC-MJX allows for high throughput training on massively parallelized environments for the rapid generation of experiment and efficient training [3]. This paper demonstrates its high throughput capabilities for simulating realistic musculoskeletal biomechanics and neural control of the mouse forelimb during a dexterous reaching task.  By recapitulating behavior observed in the lab using a musculoskeletal model of the mouse forelimb, our proposed imitation learning framework allows for us to study how the brain and body respond to physical constraints with a fully transparent simulation. We examine complex relationships of trajectories along muscle and joint manifolds during dexterous reaching behavior using nonlinear dynamical systems theory. We further utilize a nonlinear forecasting methods to examine the relationships between muscle synergies and joint dynamics in simulation and use information from simulation to predict EMG signals in vivo. 

\begin{figure}[H]
    \centering
    \includegraphics[width=0.95\linewidth]{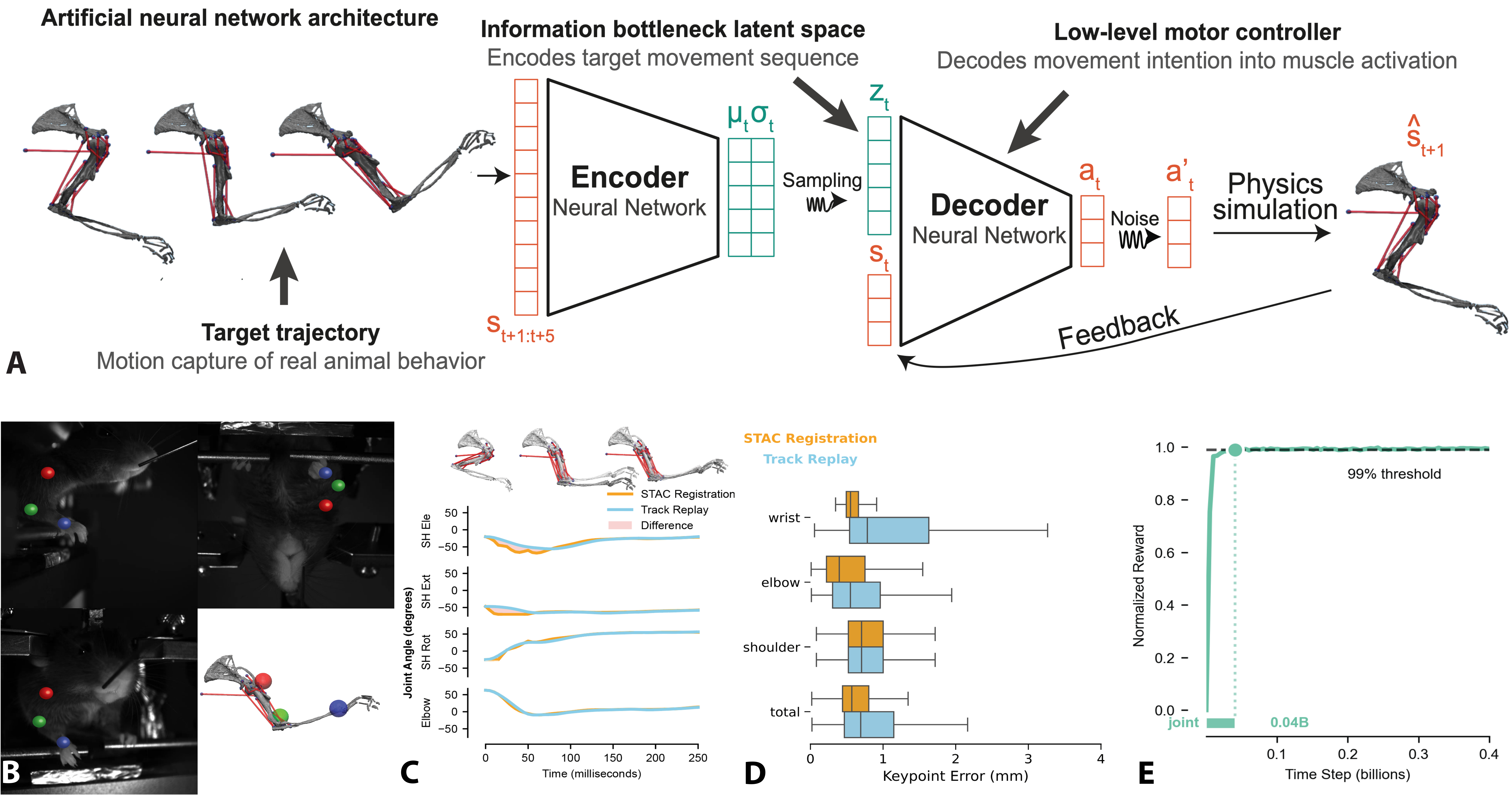}
    \caption{A. Neural network encoder-decoder architecture for imitation learning to reproduce motion capture trajectories. B. Raw video data from mouse water reaching task with 3D multi-camera pose estimation is registered to the body model using STAC-MJX. C. Kinematics of the motion capture registered model and the imitation learning performance labeled as "track replay". D. Registration error and tracking error relative to original 3D pose data. E. Joint reward throughout training. }
    \label{fig:placeholder}
\end{figure}

\FloatBarrier  

\section{Methods}
\label{gen_inst}

\subsection{Mouse forelimb water droplet reaching dataset}
Head-fixed mice (n = 4) were trained to perform goal-directed reaching movements for a water droplet in response to an auditory cue. An auditory cue was provided when the right hand was rested on a metal perch sensor for at least 0.5 sec. They were water-deprived to a maximum of 85$\%$ of their ad libitum body weight for the duration of the experiment. Each animal performed approximately 50 reaches. 

Intramuscular electromyography (EMG) electrodes were fabricated using Teflon coated 7 stranded stainless-steel wire (A-M systems, $\#$793200) in a four or six-channel configuration. Mice were anesthetized and EMG electrodes were implanted subcutaneously by inserting the attached needle into the muscle belly parallel to the longitudinal axis of the muscles of interest: biceps brachii (biceps) and triceps brachii (triceps).  

Video data was synchronously collected from 3 cameras to allow for downstream pose estimation. A subsample of this dataset was used for the purposes of this paper which contains 1 mouse (n = 1) and 46 trials as a proof of concept. The time series data was shortened to 300ms chunks aligned to the beginning of the reach. 

Procedures performed on mice were conducted according to the US National Institutes of Health guidelines for animal research and have been approved by the Institutional Animal Care and Use Committee of the Salk Institute for Biological Studies. 

\subsection{EMG signal processing}
We applied standard EMG signal processing to extract activation levels of each muscle. First, the raw EMG signals sampled at 30 kHz were band-pass filtered using a 4th-order Butterworth filter (20–1000 Hz) to remove low-frequency drift and high-frequency noise. The filtered signal was then full-wave rectified, and a low-pass filter at 50 Hz (4th-order Butterworth) was applied to extract the amplitude envelope, which estimates muscle activation level. For each trial, a fixed time window corresponding to the reach epoch was extracted, downsampled to 200 Hz by block-averaging, and clipped to a fixed length of 60 samples. The amplitude of each envelope segment was then normalized to the 98th percentile recorded across all the trials to reduce outliers and ensure comparability across trials and muscles. This procedure yields trial-aligned EMG envelopes that are suitable for direct comparison with simulated muscle activations.

\subsection{3D pose estimation with SLEAP-Anipose}
The three camera dataset was labeled using CVAT annotation tool to generate ~10,000 labeled frames per camera. This network was then run on a video consisting of 46 successful water reaching trials. Predictions of the pose of the right shoulder, elbow, and wrist were generated with a single animal pose estimation network using the U-Net architecture in the SLEAP deep learning framework. 

A calibration video was recorded by placing a chAruco board in view of all cameras. Calibration was performed using the SLEAP-Anipose package which is a wrapper around Anipose to allow for easy integration of 2D pose data from SLEAP. Calibration was performed with the SLEAP-Anipose library which uses OpenCV to perform iterative sparse-bundle adjustment. It locates the 2D points on the chAruco board and then triangulates them into 3D. Then the reprojection error is calculated by projecting back into 2D to evaluate the 3D estimation. This reprojection error is then minimized iteratively using L-BFGS to improve the 3D estimates and provide reliable triangulation. 

\subsection{Mujoco Musculoskeletal Model of the Mouse Forelimb}
A skeletal model of the mouse forelimb and estimations of muscle attachment points were provided based on light sheet microscopy data [4]. The model has 4 DoF in total: 3 DoF in the shoulder (elevation, rotation, extension) and 1 DoF in the elbow (flexion-extension). The model has 9 controllable parameters which are the following Hill-type muscle actuators: triceps (long), triceps (lateral), biceps (long), brachialis, pectoralis (clavicular), latissimus, posterior deltoid, anterior deltoid, and medial deltoid. Muscle attachment points were refined and muscle parameters were found to produce forces within the range of real mouse forelimb muscles (.2-1.2N) [4]. The in silico simulated muscle activity from the biceps long and triceps lateral were used for the subsequent comparisons with the in vivo EMG.

\subsection{Model registration with STAC-MJX}

The 3D motion capture data was registered to the Mujoco model using STAC-MJX, a parallelized framework for simultaneous tracking and calibration. Model registration consists of transforming the motion capture data into the same coordinate reference frame as the Mujoco model and then performing inverse kinematics using a Bayesian approach [6]. The registration was performed with a tolerance value of 1e-20 and max iterations of 600. After the reference clip was generated via STAC-MJX, the imitation learning model now has targets to imitate. 

\subsection{Imitation learning with Track-MJX}
The imitation learning policy was trained using \textsc{MIMIC-MJX} which is a high-speed efficient JAX implementation of imitation learning using Mujoco-MJX which allows for massive parallel execution of rollouts to gather experience for imitation learning. We use an implementation of proximal policy optimization (PPO) with an encoder-decoder architecture with an information bottleneck between the encoder and decoder in the form of a multivariate gaussian with KL regularization (Figure 1A). This latent space in the information bottleneck is what we refer to as the "motor intentions" or more specifically an encoding of the trajectory which is meant to be copied or reproduced. Initial parameter searches showed best performance with entropy cost set to .001 and KL regularization set to 1e-5. The encoder and decoder were both multi-layer perceptrons (MLPs) with three hidden layers of 512 neurons. The encoder takes in a reference observation which is a STAC registered trajectory where that trajectory length is a free parameter. 

We decompose the objective into a joint reward and complimentary cost terms to promote smooth naturalistic movements that match both the pose and muscle activations of the data observed in the lab. At time $t$, the instantaneous reward is a weighted sum
\[
r_t \;=\; \lambda_{\mathrm{joint}}\, r^{\mathrm{joint}}_t
        - \lambda_{\mathrm{ctrl}}\, c^{\mathrm{ctrl}}_t
        - \lambda_{\mathrm{energy}}\, c^{\mathrm{energy}}_t 
\]
where each reward term $r^{(\cdot)}_t \in [0,1]$ is computed using an exponentially weighted distance between the agent and the reference data.

Here is a summary of the reward and cost terms:
\begin{enumerate}
    \item \textbf{Joint Reward} ($r^{\mathrm{joint}}_t$) encourages the imitator to match the recorded joint configuration. 
Let $q_{t,i}$ denote the imitator's $i$-th joint angle at time $t$ and $\hat{q}_{t,i}$ the corresponding reference joint angle. 
The reward increases as these joint angles become closer, and is given by
\[
r^{\mathrm{joint}}_t
= \exp\!\left(
  -\,\alpha_{\mathrm{joint}}
  \sum_i \left(q_{t,i} - \hat{q}_{t,i}\right)^2
\right).
\]
    \item \textbf{Control Cost} ($r^{\mathrm{quat}}$) penalizes the sum of the square of the action $a$.
    \item \textbf{Energy Cost} ($c^{\mathrm{energy}}$) penalizes the sum over joints of the product of absolute values of joint velocities $v_t$ and actuator forces $f^{act}_t$.  \[
    c_t^{\mathrm{energy}} \;=\; 
    \sum_{j=1}^N \lvert v_{t,j} \rvert \cdot \lvert f^{\mathrm{act}}_{t,j} \rvert 
    \]

    \end{enumerate}
A grid search was performed with control cost and energy cost to find the $\lambda$ weight values for the reward and control costs which best minimized both joint distance and error between the EMG signals and action rollouts. The mean of 5 random seeds for each control cost value (0, .1, .15, .2, .25, .3 and .4) was reported, and 95\% confidence intervals were calculated across those seeds (See Figure 2A). A Fourier based metric was also used to see if the constraints reduced high frequency oscillations from 10-1000Hz. 

\subsection{Latent space and neural activity analysis}
To analyze the representational geometry of the network, we applied principal component analysis (PCA) to the mean of the intention bottleneck and to activations from each of the three hidden layers of the encoder and decoder. For each layer, activations with dimensionality $(\text{clips}, \text{timesteps}, 512)$ were reshaped into a two-dimensional matrix of shape $(\text{clips}\times \text{timesteps}, 512)$ and PCA was performed. The first three principal components were retained and the transformed data were reshaped back into $(\text{clips}, \text{timesteps}, 3)$ to preserve temporal and clip structure. The same procedure was applied to the intention representations resulting in the top three principal components. For each analysis, the proportion of variance explained by the first three components was recorded to quantify the representational compression achieved by PCA. To enable joint visualization of neural and behavioral dynamics, the three components were concatenated with the the joint angle $q_{pose}$ at that time and visualized in Figure 2E-H.

\subsection{Dynamical forecasting of EMG and simulated muscle activations}
Nonlinear simplex projection methods based on Takens' Theorem were used in order to decode EMG and simulated actions from kinematics using the pyEDM library [7,8]. Takens' Theorem establishes that partial observations of a dynamical system can be used to reconstruct generating attractor by creating a delay-embedding from the partial observation $X$. This delay embedding allows for the reconstruction of a shadow manifold $M_x$ using only the information from one variable yet is able to preserve the topology of the original attractor manifold. By creating shadow manifolds for multiple variables we can examine the correspondence of those trajectories through phase space to predict the movement of the other at a future timestep.

State-space reconstruction (SSR) methods make use of Takens’ Theorem to recover the higher-dimensional attractor of the dynamical system that generated an observed time series. Recently these methods have been extended to neural and EMG signals [9, 10, 11]. This attractor is the manifold $\mathcal{M}$ traced by the system as it evolves through state space (Fig.~6A). The state space is a Euclidean space whose axes correspond to the system’s state variables $\{x, y, z, \dots\}$.  

Takens’ theorem shows that a topology-preserving embedding of $\mathcal{M}$ can be reconstructed using delayed coordinates of a single observed variable $x(t)$:  
\[
\Phi(x_t) = \big(x(t),\, x(t+\tau),\, x(t+2\tau),\, \dots,\, x(t+(m-1)\tau)\big),
\]
where $\tau$ is the delay and $m$ the embedding dimension. This construction yields a shadow manifold $\mathcal{M}_x$, and there exists a homeomorphism (a smooth, invertible mapping) between the trajectory on $\mathcal{M}$ and its reconstruction in $\mathcal{M}_x$ (Fig.~6B,C) [3].  

Because homeomorphisms are transitive, reconstructions created from different variables are all homeomorphic to one another. SSR methods exploit this by testing for the existence of a smooth mapping $F : \mathcal{M}_x \to \mathcal{M}_y$ between manifolds reconstructed from two variables $x$ and $y$. If such a mapping exists (i.e., $\mathcal{M}_x \sim \mathcal{M}_y$), then $x$ and $y$ are likely components of the same underlying dynamical system [5].  
 
Simplex projection is a method that uses the local geometry of the reconstructed manifold for forecasting. The algorithm predicts the value of a time series $X(t)$ at a future horizon $T_p$ by identifying the nearest neighbors of the current state in the shadow manifold and interpolating their trajectories forward in time [8].

Let $E$ denote the embedding dimension, and let $k = E+1$ be the number of nearest neighbors that define the vertices of the simplex. For a point $y_t \in \mathcal{M}_x$, denote its $k$ nearest neighbors by $\{y_{N_1}, \dots, y_{N_k}\}$ with distances $d_i = \| y_t - y_{N_i} \|$. The forecast of the future value $X(t+T_p)$ is a weighted average of the future values of the time series corresponding to each nearest neighbor.

In this formulation, the simplex formed by the $k$ nearest neighbors spans a local region of the manifold, and interpolation within this region yields forecasts of the system’s future trajectory [8]. The method provides not only predictive power but also information about the intrinsic dimensionality and nonlinear dynamics. Simplex projection performance was evaluated using Spearman's $\rho$ between the observed time series and the predicted time series that we will refer to as Simplex $\rho$.  We evaluated whether the simulated joint angles could be used to decode simulated muscle activations biceps and triceps. We also evaluated whether reference joint angles and simulated actions could decode real EMG signals from the biceps and triceps.

\subsection{Dataset variables and information}
The imitation pipeline and subsequent analysis pipelines gives rise to a number of specific data types and variables which would benefit from standardization. See the Data Table 1 in the Supplementary Materials for more information about the description of all the data used to perform these analyses. See Data Table 2 for the exact configurations used to train the imitation learning models. 

\section{Results}
\label{headings}

\begin{figure}[htbp]
    \centering
    \includegraphics[width=1.0\linewidth]{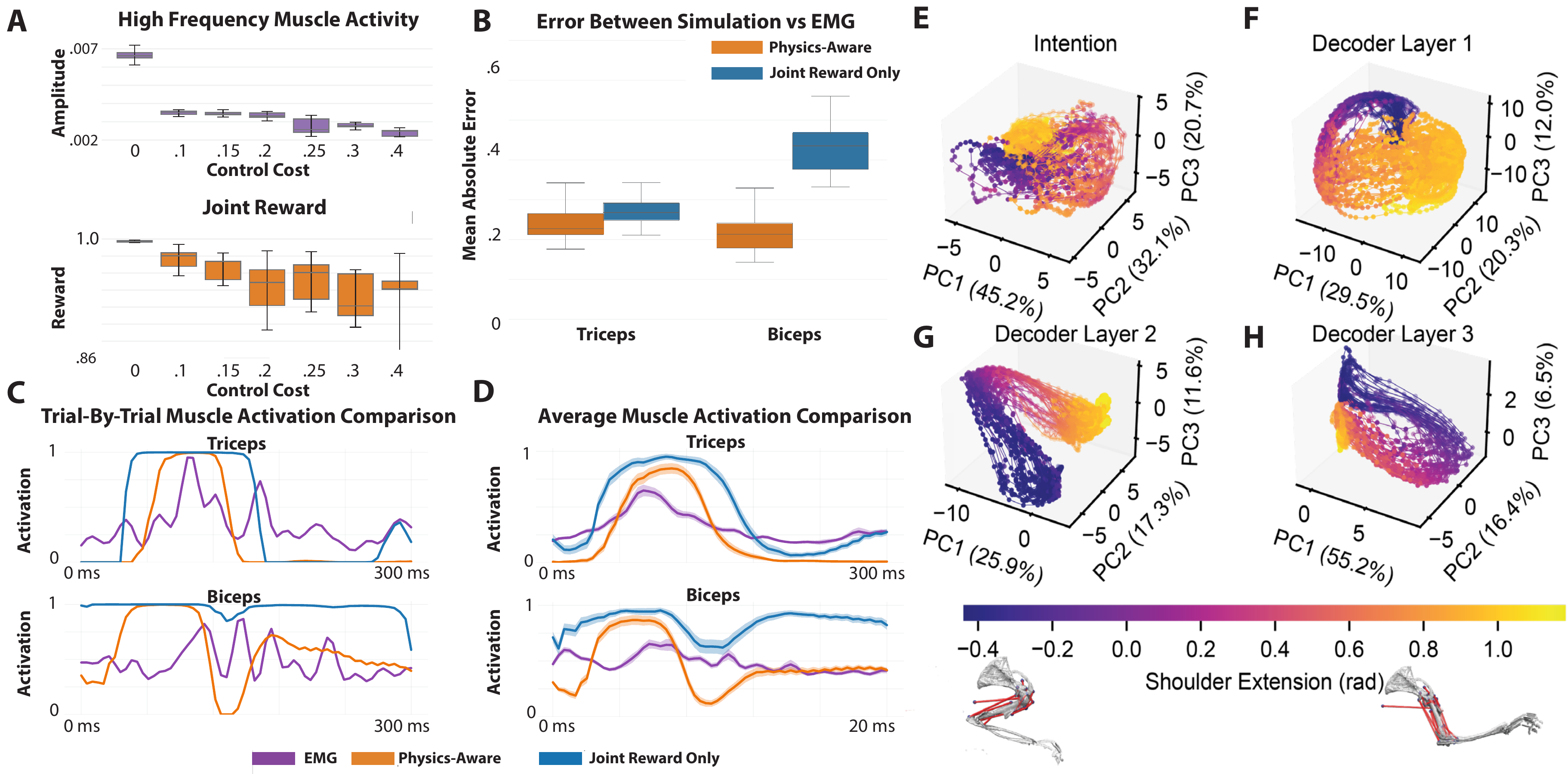}
    \caption{ A. Joint reward and high frequency activity by control cost parameter sweep, each point reflects the mean value across five independent random seeds, and the error bars denote the 95\% confidence interval computed across those seeds. B. Mean absolute error between EMG and simulated muscle activity for joint reward only and physics aware models. C. Trial-by-Trial muscle activation comparison between EMG, joint reward only and physics-aware constraints. D. Average muscle activation over time for EMG, joint reward only and physics-aware constraints with standard error of the mean in shaded region. E-H. 3D PCA embeddings of mouse arm reaching trajectories, using representations from the intention space (E) and sequential decoder layers (F-H). The axes show \% variance explained for each PC. The colormap visualizes the extent of the shoulder extension.}
    \label{fig:placeholder}
\end{figure}

\subsection{STAC-MJX registration successfully generated target movement kinematics}
The 3D mocap data was fit to the Mujoco physics model using the simultaneous tracking and calibration algorithm implemented in JAX known as STAC-MJX. This method successfully generated target movement kinematics from observed behavior in MuJoco physics environment as the model registration error between the reference pose and the imitated pose (track replay) was found to be below 1 mm on average for all keypoints (Figure 1D). 

\subsection{Imitation learning benchmarking}
The imitation learning pipeline is showing 1.2 million training steps per second on two A40s with 4096 parallel environments and 600,000 training steps per second on one A40 on 2048 parallel environments. 

\subsection{Imitation learning accurately reproduces observed joint kinematics}
The time series of reference joint angles and imitated joint angles reveals a close match in imitated kinematics (Figure 1C). The imitation learning performance (track replay) shows that in absence of physics aware constraints it can recapitulate the reference kinematics with high precision of less than 1 mm on average (Figure 1D). The model learns to maximize joint reward quickly within 40 million time steps (See Figure 1E).  See \href{https://youtu.be/t2uQnJqCkKQ?si=0tG_Hc8PwV9Dnss0}{Supplementary Video 1} to visualize imitation performance side by side with registration and the original video.

\subsection{Latent space and neural activity analysis}
The 3D visualization of the top 3 PCs were visualized for the latent bottleneck and artificial neural activations (Figure 2E-H). The latent intention layer is already tightly compressed (Figure 2E), with the first three PCs capturing 98\% of the variance (45.2\%, 32.1\%, 20.7\%). In the early decoder layers this structure loosens, with variance spreading more evenly across components (29.5\%/20.3\%/12.0\% and 25.9\%/17.3\%/11.6\% for layers 1 and 2), indicating a transient expansion that mixes and transforms task-relevant features (Figure 2F-G). By the final layer the representation reconcentrates, with the first three PCs again dominating (55.2\%, 16.4\%, 6.5\%, 78.1\% total), consistent with the network collapsing intermediate features back into a low-dimensional structure that aligns with coordinated muscle activation patterns (Figure 2H).

\subsection{Physics-aware constraints better recapitulate trial-by-trial EMG}
Adding control cost lead to a better fit between simulated and reference EMG for the biceps, there does not seem to be much of an effect for the triceps (Figure 2B). High control cost caused a reduction in joint reward and high frequency muscle activity (Figure 2A). There appears to be a general trend that as the model better fits the EMG it may lead to decreased performance on the joint reward (Figure 2A). Energy cost was not found to impact the reward or EMG fit. Trial-by-trial muscle activity was visualized simulated with activations alongside recorded EMG for both the joint-only model and the physics-aware model with control and energy constraints (Figure 2C). The average simulated muscle activity and EMG over time was plotted with standard error shown in the shaded region across all 46 clips (Figure 2D). This shows that the biceps in the joint reward only case shows high unrestrained activity in the absence of control or energy cost. When those constraints are added it leads to more conservative actuation of the muscle which appears more like the actual EMG signal. 

\subsection{Empirical dynamic modeling of simulated kinematics and muscle activity}

The simulated actions, joints and EMG data showed maximum prediction accuracy with a $\tau$ of -1. The average embedding dimension that maximized prediction accuracy of the simulated actions was E = 3. The embedding dimension for all joints was E = 2. The prediction horizon that maximized prediction accuracy was Tp = 5. For the results of the parameter search for E, $tau$, and Tp see Supplemental Figure 4.

The results show that Simplex projections of the joint angles over time can be used to decode the simulated actions, with simplex $\rho$ at .802 for Biceps. Also joint angles can be used to decode simulated muscle activations. Results also show that the reference joint angles and inferred actions can allow for the prediction of in vivo EMG signals in the triceps and to a lesser degree the biceps. 

\begin{figure}[htbp]
    \centering
    \includegraphics[width=1.0\linewidth]{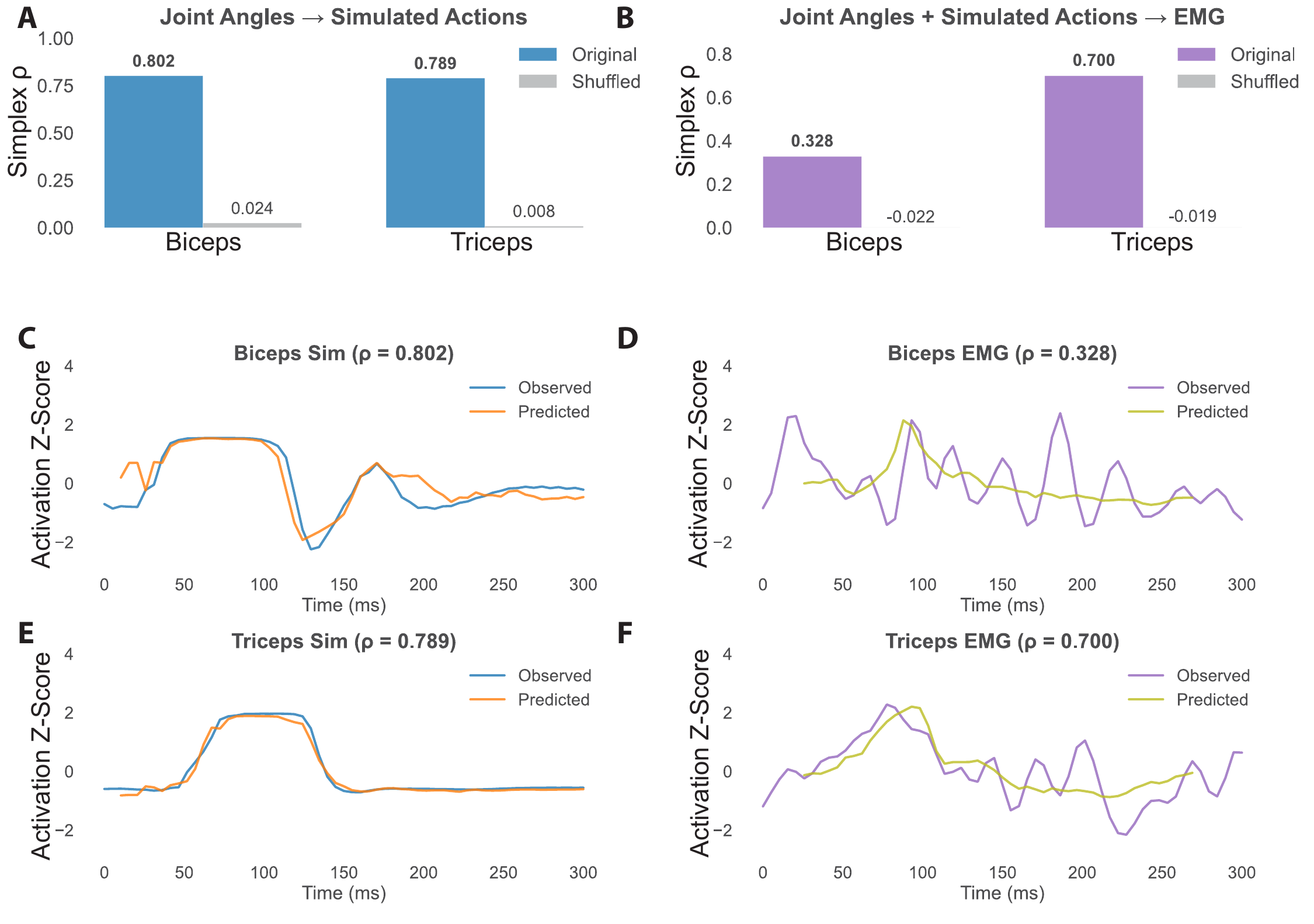}
    \caption{A. Prediction performance (simplex $\rho$) using joint angles to predict simulated biceps and triceps activations. B. Prediction performance (simplex $\rho$) using reference joint angles and simulated actions to predict Real EMG signals. C. Predictions and observed values using joint angles to predict simulated biceps activity ($\rho$ = .802). D.  Predictions and observed values using joint angles and simulated actions to predict biceps EMG ($\rho$ = .328). E. Predictions and observed values using joint angles to predict simulated triceps activity ($\rho$ = .789). F.  Predictions and observed values using joint angles and simulated actions to predict triceps EMG ($\rho$ = .7).}
    \label{fig:placeholder}
\end{figure}

\section{Discussion}
This is a demonstration of the efficiency and efficacy of both the new MIMIC-MJX pipeline and the musculoskeletal model of the mosue forelimb. 

The parameter search suggests that by adding constraints on the action space, the simulated muscle activations better fit EMG observations in the lab. Across the sweep of control and energy cost weights, we observe that increasing either the control-magnitude cost ($\lambda_{\mathrm{ctrl}}$) reliably reduces the joint reward and \emph{increases} the mean absolute error (MAE) between simulated actions and recorded EMG. This pattern indicates that, in our current setting, physics-aware regularization primarily acts as a smoothing prior that biases the policy toward lower-amplitude, lower-variability actuation. Consistent with this interpretation, control cost also produces a marked reduction in high-frequency power in the action space, yielding smoother trajectories. The energy cost did not improve the fit to the EMG data. The current definition of energy cost may be improved upon by putting a penalty on work itself rather than velocity multiplied by force.

One limitation of the dataset is that there is only one target location. The forecasting method is able to predict based on regularities in the history of the reaching trajectories and muscle activity, so if there is a lot of repetition of the same movement that may lead to increased predictability. It is known, however, that multiple different muscle synergies can lead to the same kinematics and this method appears to perform well despite this underdetermination. Future work will use multiple target locations and introduce physical and neural perturbations in order to tease out any spurious correlations which are a product of the dataset with one reach location. 

Overall, there is a narrow regime where moderate smoothing yields stable, energetically conservative behavior without severely degrading joint reward. In practice, this corresponds to operating near the knee of the frontier—choosing $\lambda_{\mathrm{ctrl}}$ large enough to remove pathological chatter but small enough to preserve transient structure required for accurate joint tracking and EMG prediction. More physiologically grounded energetic terms may retain these benefits while avoiding interference with essential bursts. Since EMG MAE is highly sensitive to these parameters, it is important to report spectral metrics, joint reward, and EMG MAE together, as improvements in one dimension can mask degradation in another. These conclusions remain preliminary given the limited number of animals, and larger datasets will be needed to assess generalizability.

The nonlinear forecasting results have shown that variables from the simulation can be utilized to effectively predict trail-by-trial EMG data observed in the lab. The forecasting method showed that the simulated actions could be reliably predicted given the joint angles. This suggests that our simulation can serve as a reliable comparison case when applying state space reconstructions. The forecasting method also showed that the triceps signal could be predicted well given the reference joint angles and simulated actions. However, the forecasting for the biceps EMG was not very effective. One possibility is that this could be due to the shoulder in the model being immobile and the biceps having to compensate with larger muscle activations than is present in a freely moving shoulder. Further data collection from more animals is required to better predict and characterize the EMG signal.

Physics-aware regularization of the action space is useful but not free: it reshapes the temporal statistics of control to be more naturalistic but can be over applied to compromise joint tracking. Practically, we recommend selecting control cost by jointly monitoring joint reward, EMG error, and high-frequency power. Methodologically, extending the objective with more physiologically grounded energetics or constraining only non–task-relevant action subspaces may preserve smoothness without suppressing informative bursts. Finally, expanding to multiple targets and controlled perturbations will help disentangle genuine muscle-level agreement from dataset regularities, providing a more rigorous test bed for policies that are simultaneously stable, energy-efficient, and neurophysiologically and biomechanically credible.

\section{Data and Code Availability}
The track-mjx repo can be found at \url{https://github.com/talmolab/track-mjx}. The stac-mjx repro can be found at \url{https://github.com/talmolab/stac-mjx}. 

Model checkpoints and rollout data can be found at \url{https://huggingface.co/talmolab/mouse-reach-mjx-neurips}. Training clip data can be found at \url{https://huggingface.co/datasets/talmolab/MIMIC-MJX/tree/main/data/mouse_arm}. EMG data, trial information and parameter search data can be found at \url{https://huggingface.co/datasets/talmolab/mouse-reach-mjx-neurips}. The figures in this paper can be replicated at \url{https://github.com/talmolab/mouse-reach-mjx-neurips}.

For more information about mimix-mjx see \url{https://mimic-mjx.talmolab.org/}

\begin{ack}
Thank you to Diego Aldarondo, Josh Merel, Mazen al Borno, Jesse Gilmer, Charles Zhang, Emil Warnberg, Bence Ölveczky, Aidan Sirbu, Elliot Abe, Adam Lee, Bing Brunton, Scott Linderman, and Blake Richards. This project was funded by NIH (5U01NS136507) to T. D. P. and E. A., L.I.F.E. Foundation to T. D. P.  and Salk Innovation Award to E. A. and T. D. P..

\end{ack}

\section*{References}

{
\small

[1] Aldarondo, D., Merel, J., Marshall, J.D., Hasenclever, L., Klibaite, U., Gellis, A., Tassa, Y., Wayne, G., Botvinick, M., \& Ölveczky, B.P. \ (2024) A virtual rodent predicts the structure of neural activity across behaviours. \ {\it Nature} {\bf 632}(8025): 594-602.

[2] Ting, L. H., \& McKay, J. L. \ (2007). Neuromechanics of muscle synergies for posture and movement.{\it Current opinion in neurobiology} {\bf 17}(6): 622-628. 

[3] Zhang, C. Y., Yang, Y., Sirbu, A., Abe, E. T. T., Wärnberg, E., Leonardis, E. J., Aldarondo, D. E., Lee, A., Prasad, A., Foat, J., Bian, K., Park, J., Bhatt, R., Saunders, H., Nagamori, A., Thanawalla, A. R., Huang, K. W., Plum, F., Beck, H. K., Flavell, S. W., Labonte, D., Richards, B. A., Brunton, B. W., Azim, E., Ölveczky, B. P., \& Pereira, T. D. (2025). MIMIC-MJX: Neuromechanical Emulation of Animal Behavior. {\it arXiv preprint} arXiv:2511.20532.

[4] Gilmer, J. I., Coltman, S. K., Cuenu, G., Hutchinson, J. R., Huber, D., Person, A. L., Al Borno, M. (2025). A novel biomechanical model of the proximal mouse forelimb predicts muscle activity in optimal control simulations of reaching movements. {\it Journal of Neurophysiology} {\bf 133}: 1266–1278. 

[5] Ramalingasetty, S. T., Danner, S. M., Arreguit, J., Markin, S. N., Rodarie, D., Kathe, C., ... \& Ijspeert, A. J. (2021). A whole-body musculoskeletal model of the mouse. IEEE Access, 9, 163861-163881.

[6] Wu, T., Tassa, Y., Kumar, V., Movellan, J.,\& Todorov, E. (2013, October). STAC: Simultaneous tracking and calibration. In 2013 13th IEEE-RAS International Conference on Humanoid Robots (Humanoids) (pp. 469-476). IEEE.

[7] Takens, F.\ (1980) Detecting strange attractors in turbulence. \ In {\it Dynamical Systems and Turbulence, Warwick 1980}:366-381.

[8] Sugihara, G. \ \& May, R.\ (1990) Nonlinear forecasting as a way of distinguishing chaos from measurement error in time series. {\it Nature} {\bf 334}:734-741.

[9] Camassa, A., Park, J., Wagner, M., Sejnowski, T., $\&$ Pao, G. M.\ (2024) Multivariate Prediction of Human Behavior in Task fMRI. In  {\it NeurIPS 2024 Workshop on Behavioral Machine Learning}.

[10] Breston, L., Leonardis, E. J., Quinn, L. K., Tolston, M., Wiles, J., $\&$ Chiba, A. A. \ (2021). Convergent cross sorting for estimating dynamic coupling. {\it Scientific Reports} {\bf 11} (1):20374

[11] Chen, L., Liu, X., Xuan, B., Zhang, J., Liu, Z., $\&$ Zhang, Y.  (2021). Selection of EMG Sensors Based on Motion Coordinated Analysis. {\it Sensors}, {\bf 21}(4), 1147.

}

\newpage
\section{Supplemental Materials}

\begin{table}[t!htbp]
  \caption{Data types and formats for rollouts}
  \label{tab:rollout-data}
  \centering
  \begin{tabular}{l p{0.35\textwidth} l l}
    \hline
    Name & Description & Format & Size \\
    \hline
    Multicamera Video         & frames $\times$ 3  
                & .mp4        & $\sim$1.5\,GB \\
    2D Pose Estimation        & frames $\times$ nodes $\times$ 2                 & .h5, .slp   & $\sim$12\,MB  \\
    Multi-Camera Calibration  & Intrinsics, extrinsics, distortions              & .toml       & $\sim$1\,KB   \\
    3D Pose Estimation        & frames $\times$ nodes $\times$ 3                 & .h5         & $\sim$8\,MB   \\
    Inverse Kinematics        & Joint angles over time from STAC-MJX                          & .h5         & $\sim$2\,MB   \\
    Registration Offsets      & Offsets between pose estimation and MuJoCo model & .h5         & $\sim$10\,MB  \\
    EMG Observations          & Biceps and triceps recorded in the lab over time & .csv        & $\sim$128\,MB   \\
    Trial info & Indexes for trial start and reach start. & .csv & $\sim$1\,KB \\
    \hline
    \multicolumn{2}{l}{\textbf{Rollouts (single file; multiple keys)}} & .h5 & $\sim$32\,MB \\
    \quad Latent Activations         & 4 latent dimensions over time           \\
    \quad Encoder Activations        & 3 layers of 512 activations over time  \\
    \quad Decoder Activations        & 3 layers of 512 activations over time   \\
    \quad Simulated Muscle Activations & 9 muscle actuators over time          \\
    \quad Simulated Kinematics       & Four joints over time                  \\
    \hline
  \end{tabular}
\end{table}

\begin{table}[h]
\hspace{0.5em}
\makebox[0pt][l]{%
\begin{minipage}{.6\linewidth}
\centering
\resizebox{1.2\linewidth}{!}{%
\begin{tabular}{lcc}
\hline
\textbf{PPO Params} & \textbf{Joint Only} & \textbf{Physics Aware} \\
\hline
num envs & 4096 & 4096 \\
batch size & 1024 & 1024 \\
num minibatches & 8 & 8 \\
learning rate & 1.00E-04 & 1.00E-04 \\
clipping epsilon & 0.2 & 0.2 \\
discounting & 0.95 & 0.95 \\
entropy cost & 0.001 & 0.001 \\
unroll length & 20 & 20 \\
kl weight & 0.00001 & 0.00001 \\
\hline
\textbf{Network Params} & & \\
\hline
encoder layer sizes & [512,512,512] & [512,512,512] \\
decoder layer sizes & [512,512,512] & [512,512,512] \\
critic layer sizes & [512,512,512] & [512,512,512] \\
intention size & 4 & 4 \\
\hline
\textbf{Sim Params} & & \\
\hline
sim dt & 0.00125 & 0.00125 \\
ctrl dt (sim\_dt * steps...) & 0.0025 & 0.0025 \\
solver & CG & CG \\
iterations & 6 & 6 \\
ls iterations & 6 & 6 \\
\hline
\textbf{Reward Params} & & \\
\hline
pos exp scale & 0.0 & 0.0 \\
quat exp scale & 0.0 & 0.0 \\
joint exp scale & 0.2 & 0.2 \\
end eff exp scale & 0.0 & 0.0 \\
body pos exp scale & 0.0 & 0.0 \\
joint vel exp scale & 0.0 & 0.0 \\
pos weight & 0.0 & 0.0 \\
quat weight & 0.0 & 0.0 \\
joint weight & 5 & 5 \\
end eff weight & 0.0 & 0.0 \\
body pos weight & 0.0 & 0.0 \\
joint vel weight & 0.0 & 0.0 \\
control cost & 0.0 & 0.15 \\
control difference cost & 0.0 & 0.0 \\
energy cost & 0.0 & 0.01 \\
variance cost & 0.0 & 0.0 \\
variance window & 0.0 & 0.0 \\
\hline
\end{tabular}}
\caption{Training configuration parameters for Joint-Only and Physics-Aware mouse arm models.}
\label{table2}
\end{minipage}%
}
\end{table}

\begin{figure}[htbp]
    \centering
    \includegraphics[width=1.0\linewidth]{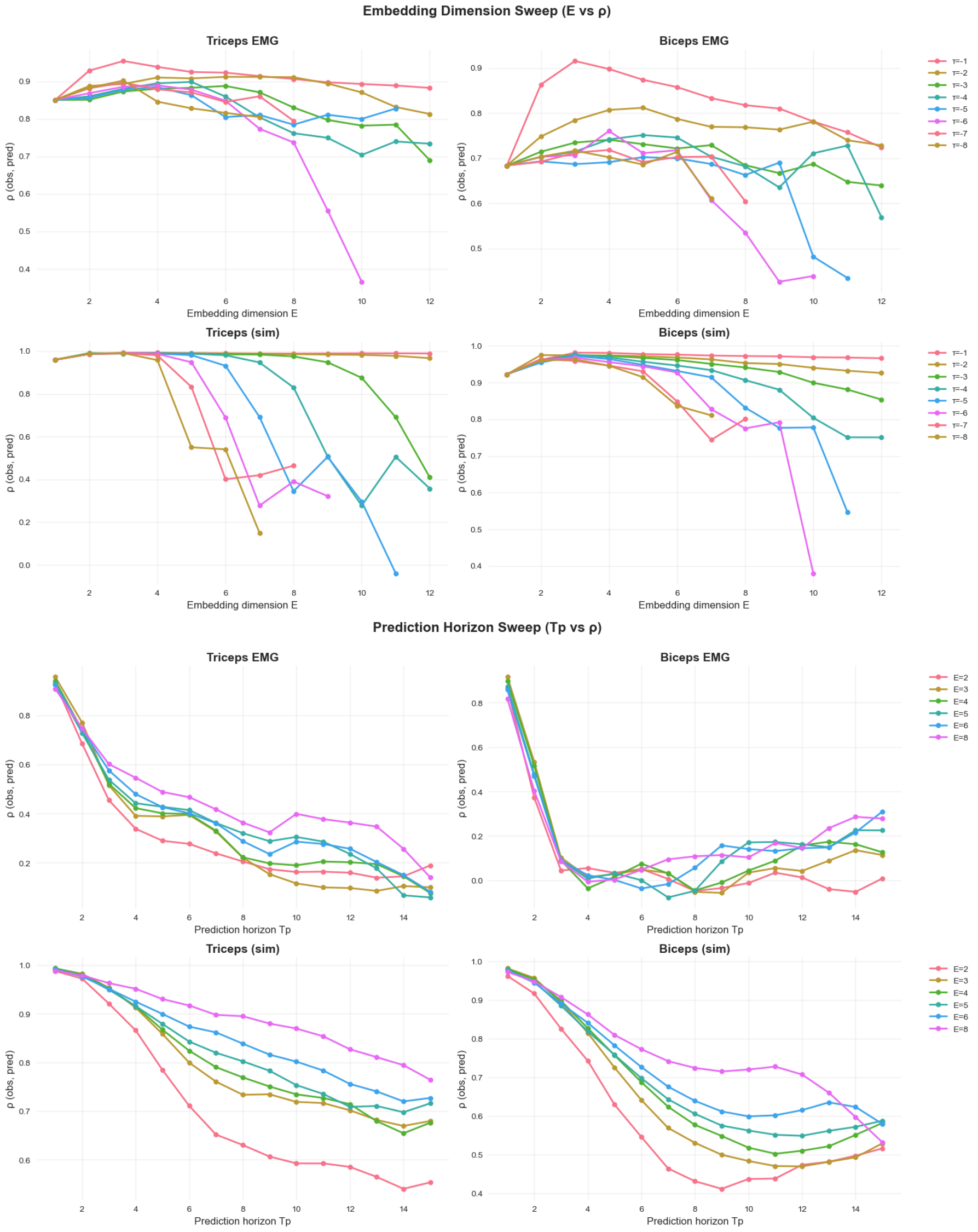}
    \caption{PyEDM parameter search. Embedding dimension and tau sweep with performance measured by simplex rho for the simulated muscle activity and EMG signals find an optimal tau of -1 and optimal embedding dimension of 3. Prediction horizon search for the simulated muscle activity and EMG signals show the presence of a nonlinearity.}
    \label{fig:placeholder}
\end{figure}

\end{document}